\documentclass[letterpaper, 10 pt, conference]{ieeeconf}  

\IEEEoverridecommandlockouts                              

\usepackage{epstopdf}
\usepackage{epsfig} 
\usepackage{amsmath} 
\usepackage{amssymb}  
\usepackage{tabularx}
\usepackage[utf8]{inputenc}
\usepackage[T1]{fontenc}
\usepackage{textcomp}
\usepackage{gensymb}
\usepackage{multirow}

\usepackage{xcolor}
\usepackage{graphicx}
\graphicspath{ {./images/}}
\usepackage{subcaption}
\usepackage{float}
\usepackage{capt-of}
\usepackage{etoolbox}
\usepackage{adjustbox}
\usepackage[colorlinks=true]{hyperref}
\usepackage{subcaption} 
\usepackage[font=small]{caption}

\usepackage{cite}

\title{\textbf{INFER: INtermediate representations for FuturE pRediction}}

\author{Shashank Srikanth$^{1,*}$, Junaid Ahmed Ansari$^{1}$, R. Karnik Ram$^{1}$, Sarthak Sharma$^{1}$, \\
J. Krishna Murthy$^{2}$, and K. Madhava Krishna$^{1}$
\thanks{$^1$Robotics Research Center, KCIS, IIIT Hyderabad, India} 
\thanks{$^*$Correspondence: {\tt\small \{s.shashank2401\}@gmail.com}}
\thanks{$^2$Mila - Quebec AI Institute, Montreal, Canada.}
\thanks{$^3$More qualitative and quantitative results, along with code and data can be found at \url{https://rebrand.ly/INFER-results}}
\thanks{The authors thank Vibhav Vineet (Microsoft Research, Redmond) and Ondrej Miksik (University of Oxford) for valuable discussions.}
}

\begin{document}

\makeatletter
\let\@oldmaketitle\@maketitle
\renewcommand{\@maketitle}{\@oldmaketitle
\centering
\includegraphics[width=17.5cm,height=6.8cm]{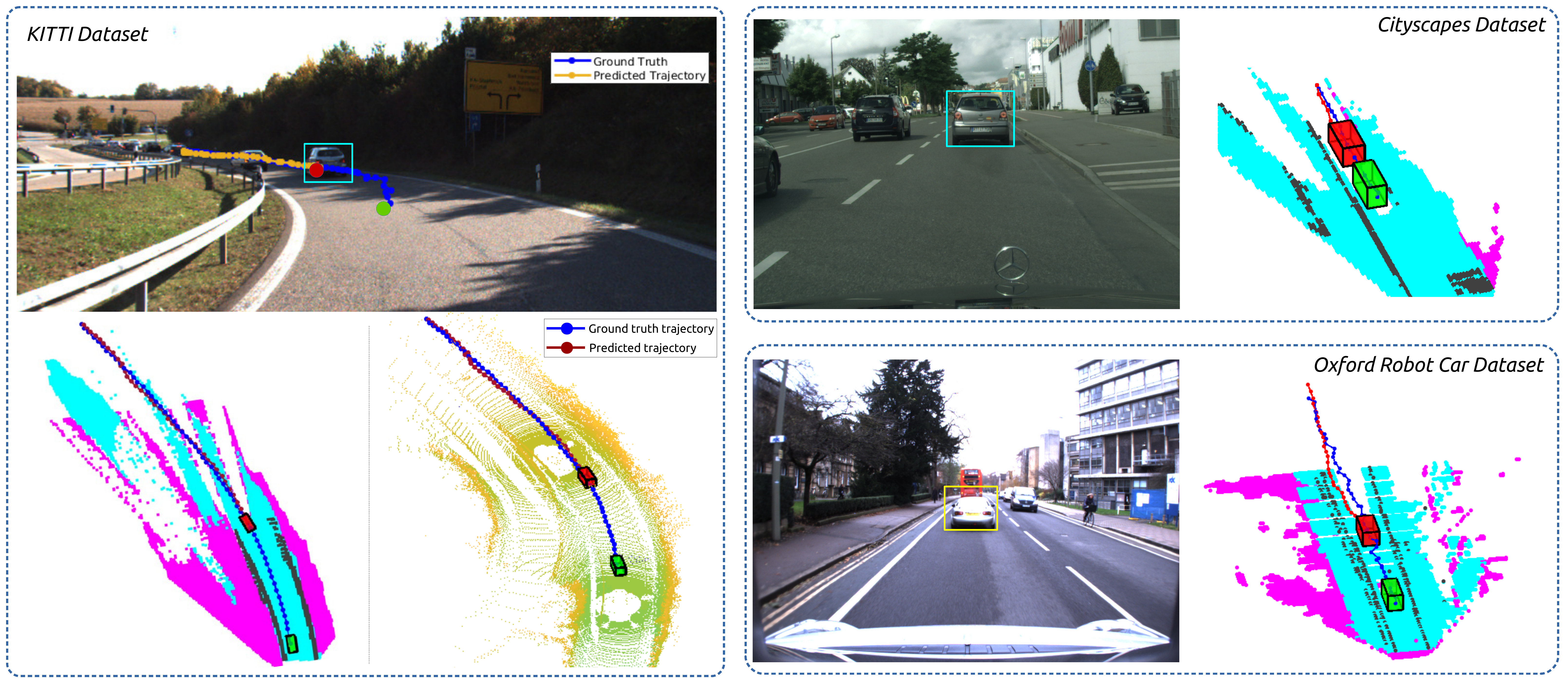}
\vspace{-0.1cm}
\captionof{figure}{\textbf{Future prediction}: \emph{Left:} (top) An interesting scenario from the KITTI dataset (LiDAR and stereo cameras used) \cite{KITTI} where the car (cyan bounding box) will turn left over the next $4$ seconds. We propose INFER (INtermediate representations for FuturE prediction): we predict plausible future trajectories, given a small set of samples from the past (the trail behind the car). The most-confident prediction of the future trajectory (orange) and the ground truth trajectory (blue) are projected onto the image for visualization. \emph{Right:} \textbf{Zero-shot transfer} to the Cityscapes \cite{cityscapes} (using stereo vision only) and Oxford RobotCar \cite{oxfordCar} (using LiDAR only). INFER tranfers across sensor modalities and driving scenarios (eg. left-lane driving in the Oxford RobotCar \cite{oxfordCar} dataset vs right-lane driving in the KITTI \cite{KITTI} and Cityscapes datasets \cite{cityscapes}. The predicted and ground-truth trajectories in 3D for each of the datasets are shown below/to the right of the image in each block. The green 3D bounding box depicts the first sighting of the vehicle of interest, and the red 3D bounding box indicates the instant from which we start predicting the future trajectory. We also visualize the LiDAR/depth information (road-cyan, lane-dark gray, and road-magenta) to demonstrate the accuracy of our predictions. A supplementary video is available at \url{https://youtu.be/wDM8EmnzLWI}}}
\label{fig:teaser_new}
\makeatother

\maketitle

\vspace{-0.5cm}
\begin{abstract}
In urban driving scenarios, forecasting future trajectories of surrounding vehicles is of paramount importance. While several approaches for the problem have been proposed, the best-performing ones tend to require extremely detailed input representations (eg. image sequences). But, such methods do not \emph{generalize} to datasets they have not been trained on. We propose \emph{intermediate representations} that are particularly well-suited for future prediction. As opposed to using \emph{texture} (color) information, we rely on \emph{semantics} and train an autoregressive model to accurately predict future trajectories of traffic participants (vehicles) (see fig. above). We demonstrate that using semantics provides a significant boost over techniques that operate over raw pixel intensities/disparities. Uncharacteristic of state-of-the-art approaches, our representations and models generalize to completely different datasets, collected across several cities, and also across countries where people drive on opposite sides of the road (left-handed vs right-handed driving). Additionally, we demonstrate an application of our approach in multi-object tracking (data association). To foster further research in transferrable representations and ensure reproducibility, we release all our code and data. $^{3}$
\end{abstract}



\section{Introduction}
\label{sec:introduction}

Deep learning methods have ushered in a new era for computer vision and robotics. With very accurate methods for object detection \cite{MaskRCNN} and semantic segmentation \cite{DenseRelationNet}, we are now at a juncture where we can envisage the application of these techniques to perform higher-order understanding. One such application which we consider in this work, is predicting future states of traffic participants in urban driving scenarios. Specifically, we argue that constructing \emph{intermediate representations} of the world using off-the-shelf computer vision models for semantic segmentation and object detection, we can train models that account for the multi-modality of future states, and at the same time transfer well across different train and test distributions (datasets).

Our approach, dubbed \emph{INFER (INtermediate representations for distant FuturE pRediction)}, involves training an autoregressive model that takes in an \emph{intermediate representation} of past states of the world, and predicts a multimodal distribution over plausible future states. The model takes as input an intermediate representation of the scene semantics (intermediate, because it is neither too primitive (eg. raw pixel intensities) nor too abstract (eg. velocities, steering angles). Using these intermediate representations, we predict the plausible future locations of the \emph{Vehicles of Interest} (VoI). 

Several state-of-the-art approaches for the future prediction task have been proposed over the last couple of years \cite{desire,seq2seq_prediction,conv_social_pooling,sophie}. However, most such approaches are well-tuned to the data distribution they have been trained to match. To perform well on a different dataset (eg. from a different city/country), such models need to be retrained/finetuned. We postulate that this need for additional training is due to the mismatch between the source (train) and target (test) distributions (datasets). This problem is further aggravated when weather conditions or scene layouts change, as would happen due to climatic variation or a switch from left-handed to right-handed driving in certain countries. In this work, we argue that, the difficulty in transferring across different target distributions is by virtue of the input data being primitive (eg. raw pixel intensities \cite{desire,sophie}). 

Motivated by this, we use \emph{semantics} as an intermediate-representation and train a neural autoregressive model that generalizes (\emph{zero-shot}) to datasets it has never been trained on. Specifically, we train a model that predicts future trajectories of traffic participants over the KITTI \cite{KITTI} dataset, and test it on different datasets \cite{oxfordCar,cityscapes} and show that the network performs well across these datasets which differ in scene layout, weather conditions, and also generalizes well across sensing modalities (models trained using a combination of stereo and LiDAR data perform well even when either of those modalities are absent at test time). In addition to transferring well, we outperform the current best future prediction model \cite{desire} on the KITTI\cite{KITTI} dataset while predicting deep into the future ($3-4$ sec) by a significant margin. Furthermore, we conduct a thorough ablation study of our intermediate representations, to answer the question ``\emph{What kind of semantic information is crucial to accurate future prediction?}". We then showcase one important application of future prediction---multi-object tracking---and present results on select sequences from the KITTI \cite{KITTI} and Cityscapes \cite{cityscapes} datasets.

In summary, we make the following contributions:
\begin{enumerate}
    \item We propose \emph{INFER}, an autoregressive model to forecast future trajectories of dynamic traffic participants (vehicles). We beat multiple challenging baselines, as well as current state-of-the-art approaches while predicting future locations of vehicles deep into the future by a significant margin.
    \item Uncharacteristic of prior art, \emph{INFER} transfers \emph{zero-shot} to datasets it has never been trained on, whilst maintaining similar performance. We show results of zero-shot transfer on the Cityscapes \cite{cityscapes} and Oxford RobotCar \cite{oxfordCar} datasets, using a model trained on sequences from KITTI \cite{KITTI}.
    \item We carry out principled ablation studies to gather empirical evidence to answer the question ``\emph{What kind of semantics generalize across datasets?}". We also carry out an ablation study on how the model performs on different frame rates than the one(s) it was trained on.
    \item We make publicly available a cross-dataset benchmark for future prediction, comprising augmented manual annotations and semantics for the  datasets that we evaluate on.
\end{enumerate}


\begin{figure*}[!hbt]
\begin{minipage}{\linewidth}
    \centering
    \includegraphics[width=\textwidth, height=7cm]{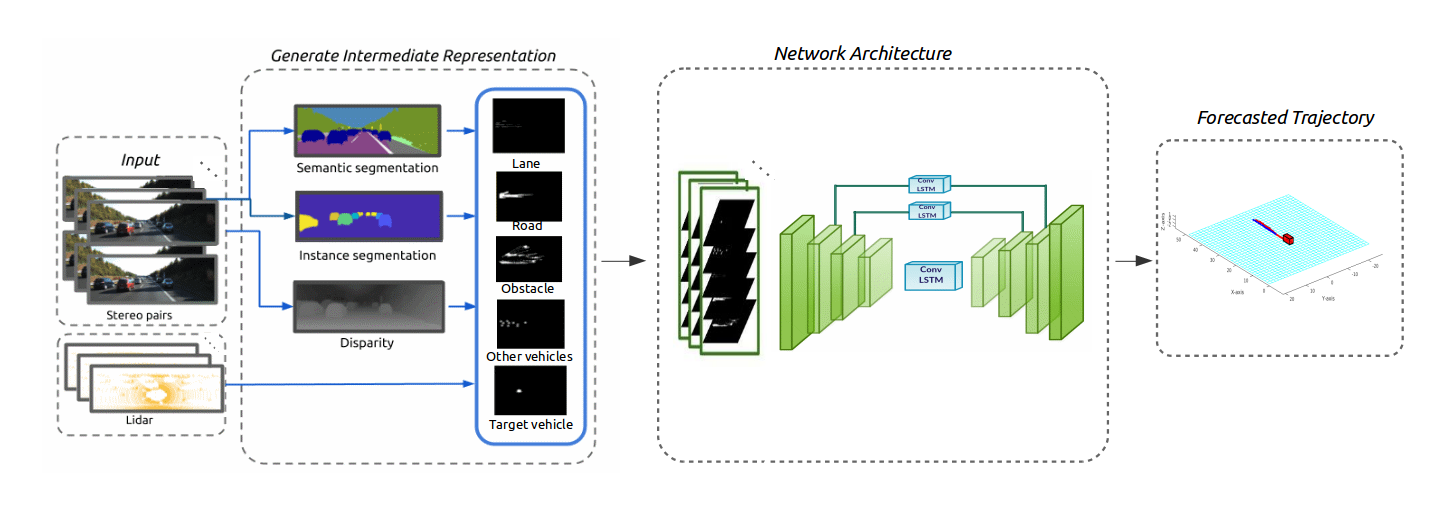}    
    \captionof{figure}{\textbf{Proposed framework}: We first \emph{generate intermediate representations} by fusing monocular images with depth information (from either stereo or Lidar), obtaining semantic  and instance segmentation from monocular image, followed by an orthographic projection to bird's-eye view. We use \cite{MaskRCNN} to detect and \cite{ioutracker} to track objects.
    The generated intermediate representations are 
    fed through the network, and finally we show the resulting prediction of the target vehicle's trajectory registered in the sensor coordinate frame.}
    \label{fig:qualkitti}
\end{minipage}
\end{figure*}

\section{Related Work} \label{relatedwork}

Of late, several approaches have been proposed to tackle the problem of future prediction in dynamic scenes. Here, we summarize a few of them, while drawing parallels and contrast to our own.

\subsubsection*{\textbf{Classical Methods}}
The problem has been studied extensively in the classical probabilistic modelling paradigm \cite{barth2009estimating},\cite{danescu2011modeling}. However, these approaches typically make strong assumptions about the scene and/or require explicit hand-modelling. In contrast, the recent more powerful learning models seem to show better promise in learning the vehicle models and in accurately forecasting their trajectories \cite{desire}.


\subsubsection*{\textbf{IRL for path prediction}} Another set of approaches involve using Inverse Reinforcement Learning to estimate the action taken by an agent at each time step and predict the future paths subsequently by applying the estimated actions sequentially at the current target location. Activity forecasting is done in \cite{kitani2012activity} by combining semantic maps with ideas from optimal control thoery. In contrast to our approach, the authors use a stationary survelliance camera to forecast activities of pedestrians, where as we predict future locations of vehicles in highly dynamic scenes. 

\subsubsection*{\textbf{RNNs for future prediction}} RNNs have been used for several sequence modelling tasks and can be used for generating sequential future prediction outputs. DESIRE \cite{desire} proposes a stochastic recurrent encoder-decoder network for predicting the future trajectories of agents in dynamic environments. The overall model comprises three components, which includes a conditional VAE followed by an RNN encoder-decoder to generate a set of diverse plausible future states, followed by an inverse optimal control-based ranking and refinement module to rank the predictions and enforce global consistency. Despite using semantic information, the authors however do not claim transfer across datasets and have different models for the Stanford Drone Dataset \cite{stanford_drone} and the KITTI dataset \cite{KITTI}. In contrast, we are able to show zero-shot transfer of our approach on different datasets like  \cite{oxfordCar,cityscapes} when trained on \cite{KITTI}.

\cite{SociallyAwareKalman} uses an interaction layer and a Kalman filter that is embeded in the architecture to learn high variance sensor inputs. They evaluate on the NGSIM dataset which consists of highway scenes. In \cite{carnet}, the authors leverage scene semantics and the past motion trajectory to predict future trajectories. They evaluate their approach on datasets recorded from a stationary camera, and show transfer across unseen scenes from the datasets, rather than cross dataset transfer as we do. The approach of \cite{virdi2017using} uses geometric and motion properties of the vehicle in the form of yaw, velocity, acceleration, and bounding box. This approach however does not leverage any scene semantics and does not show any results of transfer across different datasets. All the above approaches rely purely on the autoregressive nature of LSTMs to predict hypotheses for future trajectories of participants.



\subsubsection*{\textbf{Generative models}}  \cite{social_gan, sophie} predict pedestrian trajectory and exploit generative adversarial networks (GANs) to regularize the output future trajectories to be more \emph{realistic}. In \cite{social_gan}, a novel pooling mechanism was introduced for aggregating information across people, and socially plausible future states are predicted by training adversarially against a recurrent discriminator. 


\subsubsection*{\textbf{Manoeuvre-based approaches}}
Another set of approaches like \cite{conv_social_pooling}, \cite{Deo_2018} use manoeuvre-based LSTMs for social interaction-aware trajectory prediction. In \cite{conv_social_pooling}, a convolutional social pooling layer is proposed for robust learning of interdependencies in vehicle motion. The authors define $6$ classes of maneuvers (eg. brake, accelerate, etc.) and assign a probability to each maneuver class, to obtain multi-modal outputs. Convolutional social pooling aids the learning process by creating a \emph{social tensor} which is then fed to the decoder to infer a distribution over plausible maneuvers. They showcase results on \cite{us101,i80} by forming train and test splits comprising of sequences from both the datasets, rather than showcasing transfer across one another. 



\subsubsection*{\textbf{Occupancy grid approaches}} An approach similar to ours is \cite{Kim_2017}, in which the the future locations of the vehicle is predicted on an occupancy grid map, with results shown on a highway driving dataset. In \cite{seq2seq_prediction}, the authors propose a seq2seq model that takes as input at time $t$ the ego vehicle's velocity and yaw angle, as well as the surrounding vehicles' positions and velocities \& generates $K$ locally best trajectory candidates over an occupancy grid. The approach is devoid of scene semantics and no transfer across dataset is shown. 

Our approach, which predicts future location of the VoI deep into the future (upto 4 sec) in highly dynamic environments leverages semantics, depth information and orthographic mapping to represent the raw sensor data in the form of a novel representation that not only reifies cross-dataset transfer : from \cite{KITTI} to \cite{cityscapes,oxfordCar} but is also able to generalize well to cross-sensor modalities (Eg. from LiDAR to stero, see Sec. \ref{sec:results}). 




%

\section{Intermediate Representations}
\label{sec:intermediate_representations}

In this section, we describe the intermediate representations used by INFER to describe an urban driving scenario. Our design philosophy is based on the following three desired characteristics that knowledge representation systems must possess:
\begin{enumerate}
    \item \textbf{Representational adequacy}: Such representations must have the capability to \emph{adequately} represent task-relevant information.
    \item \textbf{Inferential adequacy}: They must also have the capability to infer traits that cannot otherwise be inferred from the original unprocessed data.
    \item \textbf{Generalizability}: These representations must necessarily generalize to other data distributions (for the same task).
\end{enumerate}

Scene representation for the task of future prediction spans a broad spectrum. On one end is solving the future prediction problem based on raw RGB input and predicting the future locations of the VoI in the image space in the form of heatmaps. On the other end is leveraging the geometric information of the VoI in the form of 3D coordinates w.r.t ego vehicle, rotational parameters, relative velocity etc and then regressing to locations of the VoI vehicle deep into the future. The former representation operates on raw RGB data without reasoning about the scene geometry in any form and hence the predicted location of the VoI in the 2D image space would again need to be interpreted in 3D. Approaches like \cite{sfmlearner} that reason about depth from single view suffer from dataset dependencies, struggling to show transfer on KITTI \cite{KITTI} after being trained on Cityscapes \cite{cityscapes}. The dependence of the network to operate on the RGB input enfeebles the transfer on datasets with different RGB pixel values and sensor modalities, as seen in DESIRE \cite{desire} where different networks were used for different datasets. 

The other end of the scene representation spectrum deals with representing the VoI in form of its geometric properties viz. 3D location w.r.t ego vehicle, rotational parameters, velocity. Although this form of representation captures the depth and geometric properties, it is not able to reason about the scene layout in any sense. Using such kind of representation makes it infeasible to reason about scene semantics like road, lanes, other vehicles or obstacles present in the scene, forcing the system to reason purely based on the relative geometry of VoI.



To this end, we choose a representation that takes the best of both worlds. The proposed representation does not rely heavily on the camera viewing angle, as camera mounting parameters (height, viewing angle, etc.) vary across datasets, and we want our approach to be robust to such variations. We hence adopt a birds-eye view as a canonical reference frame that we transform sensor data to. Further, this helps get rid of undesirable perspective distortion effects. The proposed approach dispenses the dependency on raw pixel intensities and camera intrinsic matrix by extracting semantics of the scenes. This also brings into play the crucial role of semantics that the predictions must take into account as shown in the ablation study in Table \ref{table:ablationKITTICityscapes}. We encompass scene geometry by using depth sensors to project the scene to an orthographic view. Hence, the intermediate representation reasons about the world, capturing scene layout as well as scene geometry, generalizing well not only across different datasets viz. KITTI \cite{KITTI}, Cityscapes \cite{cityscapes} and Oxford Robot Car \cite{oxfordCar} but showcasing cross-modality transfer from Lidar in KITTI \cite{KITTI} (and stereo depth) \& Oxford Robot Car \cite{oxfordCar} to stereo depth in Cityscapes\cite{cityscapes}.

The scene is represented by a five-channel occupancy grid in the birds-eye view. Each of these channels contains complementary semantic cues from the scene, namely \emph{obstacles}, \emph{road}, \emph{lane markings}, \emph{target vehicle} and \emph{other vehicles} - the intermediate scene concepts. All the five channels are in metric units and are generated from stereo image pairs and discriminative learning methods. The camera is at the center-left of the occupancy grid channels, and faces the right end of the grid with coordinate system being the cannonical camera coordinate system (i.e. the X-axis and Z-axis points towards top and right of the grid, respectively). Each generated grid is of the size $512\times512\times1$,  where each pixel has a resolution of $0.25$m.

As shown in Fig. \ref{fig:qualkitti}, we first perform semantic and instance segmentation of the left camera image (considering the left camera to be the coordinate frame of the grid) \cite{inplaceBatchNorm,MaskRCNN}, and generate the disaprity maps from the stereo image pairs using PSMNet \cite{psmnet}. 

The disparity images, segmentation masks, and the camera parameters are used to generate point clouds for static scene categories (viz. \emph{lane markings}, \emph{road} and \emph{obstacles}). These semantically classified point clouds are then projected top-down into their respective discrete and fixed size occupancy grids. The 3D points which do not project within the bounds of the occupancy grids are truncated.

Occupancy grids for vehicles are generated in a similar fashion, except that we use instance segmentation to distinguish between different vehicles. We also track these vehicles across time using an appearance and geometry based multi-object tracker \cite{ioutracker}. For each vehicle, we have one channel in the occupancy grid representing its own position - the \emph{target vehicle} channel. The \emph{other vehicles} channel represents the positions of the remaining traffic participants.

The proposed representation is a medley of the two ends of the scene representation spectrum for the task of future prediction as it captures the critical task specific information like scene layout, depth information, and semantic information. Hence, it is \emph{representationaly adequate}. The inclusion of depth, orthographic mapping and sematics adorns our representation with traits that cannot be inferred directly from the raw unprocessed data in the form of only RGB images or depth sensor. Having a fine-grained semantic map (different channels for building, kerb etc.) does not add much advanced semantic information than what is already captured by our \emph{obstacle channel} (which represents buildings, kerb, vegetation together), hence making the representation \emph{inferentialy adequate}. This representaion can be infered from any raw, unprocessed data and as we show in Section \ref{sec:results}, transfers well to a variety of datasets for the task of future prediction, hence is \emph{generalizable}.

\subsubsection*{Leveraging prior semantic maps}

One aspect of autonomous driving scenarios we wish to leverage is that using prior maps, it is possible to obtain a coarse estimate of the \emph{road} and \emph{lane} semantic channels. Specifically, we adopt the strategy in \cite{look_around} that uses OpenStreetMap (OSM) \cite{OpenStreetMap} in conjunction with GPS, and align the OSM with the current \emph{road channel}. This gives us access to estimates of the \emph{road} and \emph{lane} channels for future frames, which we demonstrate in Section \ref{sec:results} to boost performance significantly. At test time, however, note that we have no \emph{a priori} information about other vehicles on the road, as those are the very attributes we wish to predict.

\section{INFER: INtermediate representations for distant FuturE pRediction}
\label{sec:approach}

We formulate the trajectory prediction problem as a per-cell regression over an occupancy grid. We use the intermediate representations introduced in the previous section to simplify the objective and help the network generalize better. We now detail the network architecture and the training/testing procedure.

\subsection{Problem Formulation}
Assume that we are given a sequence of intermediate representations $\tau = \{\mathcal{I}_t\}_{t=1}^{M}$ for a particular VoI $\mathcal{V}$, where $\mathcal{I}_t$ denotes the intermediate representation for VoI $\mathcal{V}$ at time $t$. The objective of future state prediction is then to predict a (multi-hypothesis) distribution $\{\mathcal{F}_t\}_{t=M+1}^{N}$, where $N > M$ ($M, N \in \mathbb{Z}$). Each $\mathcal{F}_t$ is a distribution over a regular grid $(x_i, y_i)$ that represents the likelihood of VoI $\mathcal{V}$ being at $(x_i, y_i)$, conditioned on $\tau$.

\subsection{Network Architecture}
We train an autoregressive model that outputs the VoI's position on an occupancy grid, conditioned on the previous intermediate representations. We use a simple Encoder-Decoder model connected by a convolutional LSTM to learn temporal dynamics. The proposed trajectory prediction scheme takes as input a sequence of intermediate representations and produces a single channel output occupancy grid.

The input grids to the network are resized to spatial dimension of $256 \times 256$ from $512 \times 512$ to reduce the network size and training overhead. First, through a series of convolution, pooling and non-linearity operations, the Encoder reduces the input resolution from $5 \times 256 \times 256$ to $64 \times 32 \times 32$ . This reduced tensor is passed through the Convolutional LSTM sandwiched between the encoder-decoder. The LSTM consists of $64$ convolution filters each with a seperate hidden and cell state with kernel dimension $3 \times 3$. The output of the LSTM is upsampled via the decoder to a resolution of $8 \times 256 \times 256$ and then this tensor is convolved with a $1 \times 1$ filter to provide the future location of the vehicle, which is a likelihood map of dimension $1 \times 256 \times 256$.

During downsampling of input via the encoder, spatial information is lost in the pooling step. Also, as we are dealing with future prediction of possible location of a vehicle after upsampling, we need to retain the temporal information from each downsampled step in encoder. In order to capture both the spatial as well as temporal information, we add \emph{skip connections} between corresponding encoder and decoder branches. We also experiment without convolutional LSTMs over the skip connections.


\subsection{Training objective}

Our training objective comprises two terms: a reconstruction loss term, and a safety loss term. More formally, the reconstruction loss $\mathcal{L}_{rec}$ penalizes the deviation of the predicted future distribution $\hat{\mathcal{F}}$ from the actual future state $\mathcal{F}^*$. Mathematically, 
\begin{equation}
    \mathcal{L}_{rec} = \| \hat{\mathcal{F}} - \mathcal{F}^* \|^2
\end{equation}
We also add a \emph{safety} loss term, that penalizes all predicted states of vehicles that lie in an obstacle cell.
\begin{equation}
    \mathcal{L}_{safe} = \| \mathcal{O} \odot \hat{\mathcal{F}} \|
\end{equation}
Here, $\mathcal{O}$ denotes the \emph{obstacle} channel, and $\odot$ denotes the elementwise matrix product.

\subsection{Training Phase}

To train the model, we feed intermediate representations corresponding to the first $2$ second of a sequence to provide sufficient context to the convolutional LSTM. Thereafter, we operate akin to sequence-to-sequence models, i.e., we obtain an output from the network, construct an intermediate representation using this output and the next incoming frame, and feed this into the network as a subsequent input. We train the model by truncated backpropagation through time, once $20$ frames are predicted.

\subsection{Test Phase}

During the test phase, intermediate representations from the first $2$ sec is used to initialize the LSTM and the remaining frames are predicted one step at a time. The final output predicted by the network is upsampled to a resolution of $512 \times 512$ using bilinear interplotation and the point with the highest activation in the heatmap is chosen as the predicted location of the target vehicle. Each grid cell of a grid size of $512 \times 512$ correponds to $25 \times 25$ cm$^2$ area which is suitable resolution in real world for autonomous driving scenarios. 

\section{Results}
\label{sec:results}

\subsection{Dataset}
Most approaches to future trajectory prediction \cite{desire,karasev2016intent} demonstrate results over the KITTI \cite{KITTI} autonomous driving benchmark. (\cite{karasev2016intent} deals only with pedestrians). But, the KITTI benchmark alone does not address many challenges that trajectory forecasting algorithms face in real-world operation. In us humans, one would expect to learn a prediction policy in a particular city (or a toy driving environment/park) and expect it to generalize to newer scenearios, even across cities.

Hence, we expand our test dataset to comprise sequences from the KITTI \cite{KITTI}, Cityscapes \cite{cityscapes}, and Oxford RobotCar \cite{oxfordCar} datasets, for they span a number of cities, provide for enough weather variations, and also exhibit a switch from right-side (Germany) to left-side (the United Kingdom) driving.


The proposed approach is trained on KITTI dataset. We take $21$ sequences from the train split of the KITTI Tracking benchmark and 5 sequences from the KITTI Raw dataset. We use ONLY these $26$ sequences for training and validating our models. We extract a total of $223$ trajectories from these sequences, and then divide them into train and test : $178$ trajectories for train and $45$ trajectories for test, comprising of over $11$K frames. We perform a $5$ fold cross validation on these $26$ sequences. The length of these sequences vary, from a minimum of $3$ sec ($30$ frames) to a max of $6$ sec ($60$ frames, the frame rate for KITTI is $10$ fps).

To highlight the core idea of learning transferrable representations for future prediction, we test the best performing model from the above split (i.e., the KITTI dataset) on sequences from Cityscapes \cite{cityscapes} and Oxford RobotCar \cite{oxfordCar}.

Specifically in Cityscapes \cite{cityscapes}, we choose $26$ trajectories spread over $15$ different cities. The Cityscapes dataset \cite{cityscapes} provides sequences of length $30$ frames, recorded at $17$ fps, resulting in trajectories of length upto $1.76$ seconds, comprising of approx $800$ frames.

The Oxford RobotCar \cite{oxfordCar} consists of several sequences recorded over different routes through Oxford, UK. The dataset was recorded at $16$ fps and we choose a few trajectories with a duration of $4$s or more, comprising of approx $500$ frames. For both, \cite{oxfordCar} \& \cite{cityscapes} we pass every alternate frame while testing.

We evaluate all our models using the average displacement error (ADE) metric which is defined as the average L2 distance between the ground truth and predicted trajectories, over all vehicles and all time steps. We compare our methods with the folowing baselines:
\begin{itemize}

\item \textit{Markov-Baseline:} A simple discrete Bayes filter implementation over a grid that uses a constant velocity motion model\cite{thrun2005probabilistic}.

\item \textit{RNN Encoder Decoder\cite{virdi2017using}:} A RNN encoder-decoder model that regresses to future locations based on the past trajectories and vehicle information in the form of yaw, velocity, acceleration, bounding box coordinates etc.

\item \textit{DESIRE-S\cite{desire}:} The best-performing variant from \cite{desire}, that uses a scence context fusion (SCF) module.

\item \textit{ConvLSTM-Baseline:} Our ConvLSTM architecture, taking a 4 channel input, first 3 being the RGB scene in bird's eye view and the 4th channel being the target vehicle channel. That is, a variant of our model that operates on pixel intensities, as opposed to semantics.

\item \textit{INFER:} Our proposed model but with a single convolutional LSTM. 

\item \textit{INFER-Skip:} The same model as \emph{INFER} but with $2$ additional convolutional LSTMs which serve as skip connections. 

\end{itemize}

\begin{figure}[!t]
    \begin{subfigure}{0.5\linewidth}
        \centering
        \includegraphics[scale=0.26]{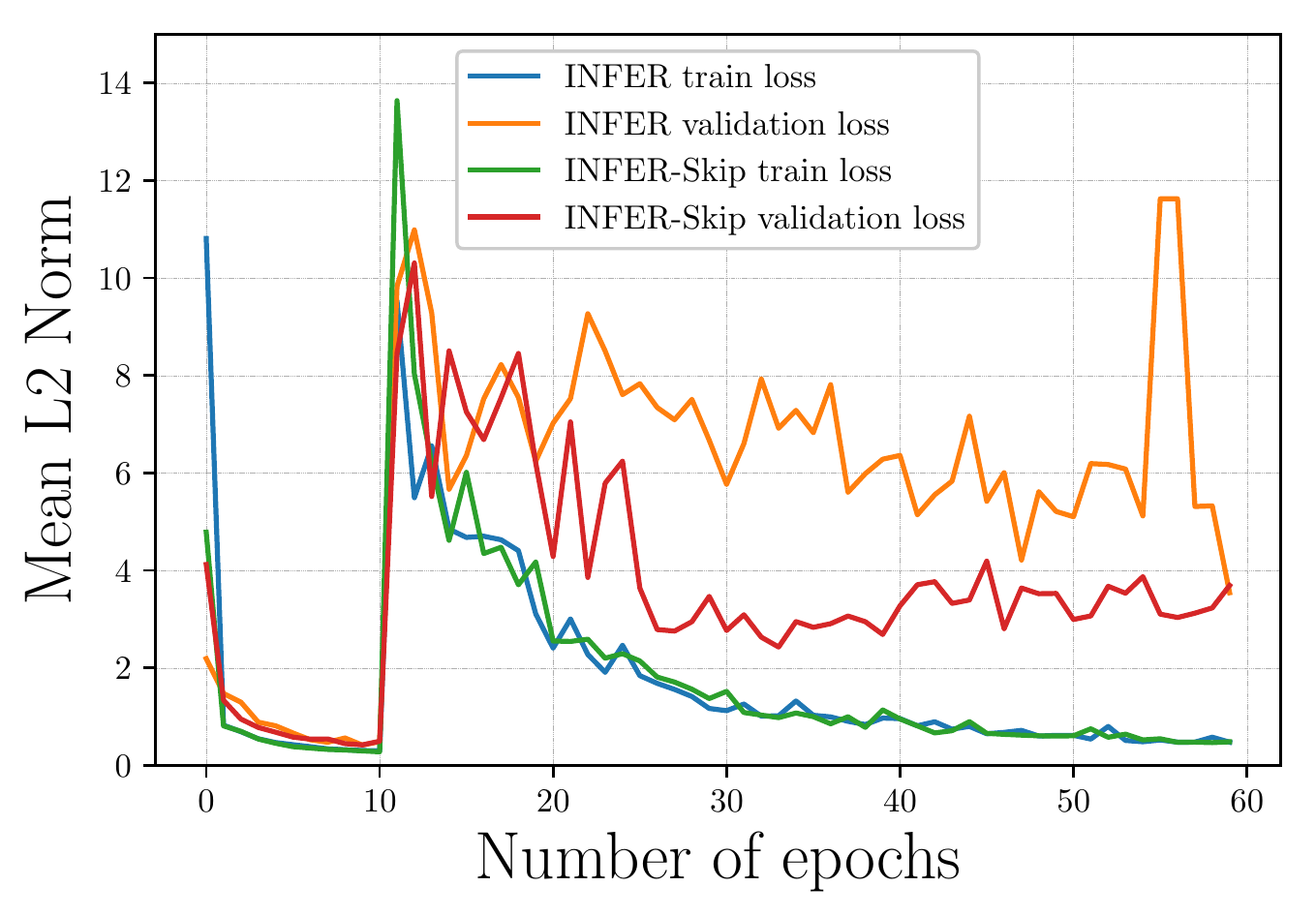}
        \caption{}
        \label{fig:network_loss}
    \end{subfigure}%
    \begin{subfigure}{0.5\linewidth}
        \centering
        \includegraphics[scale=0.22]{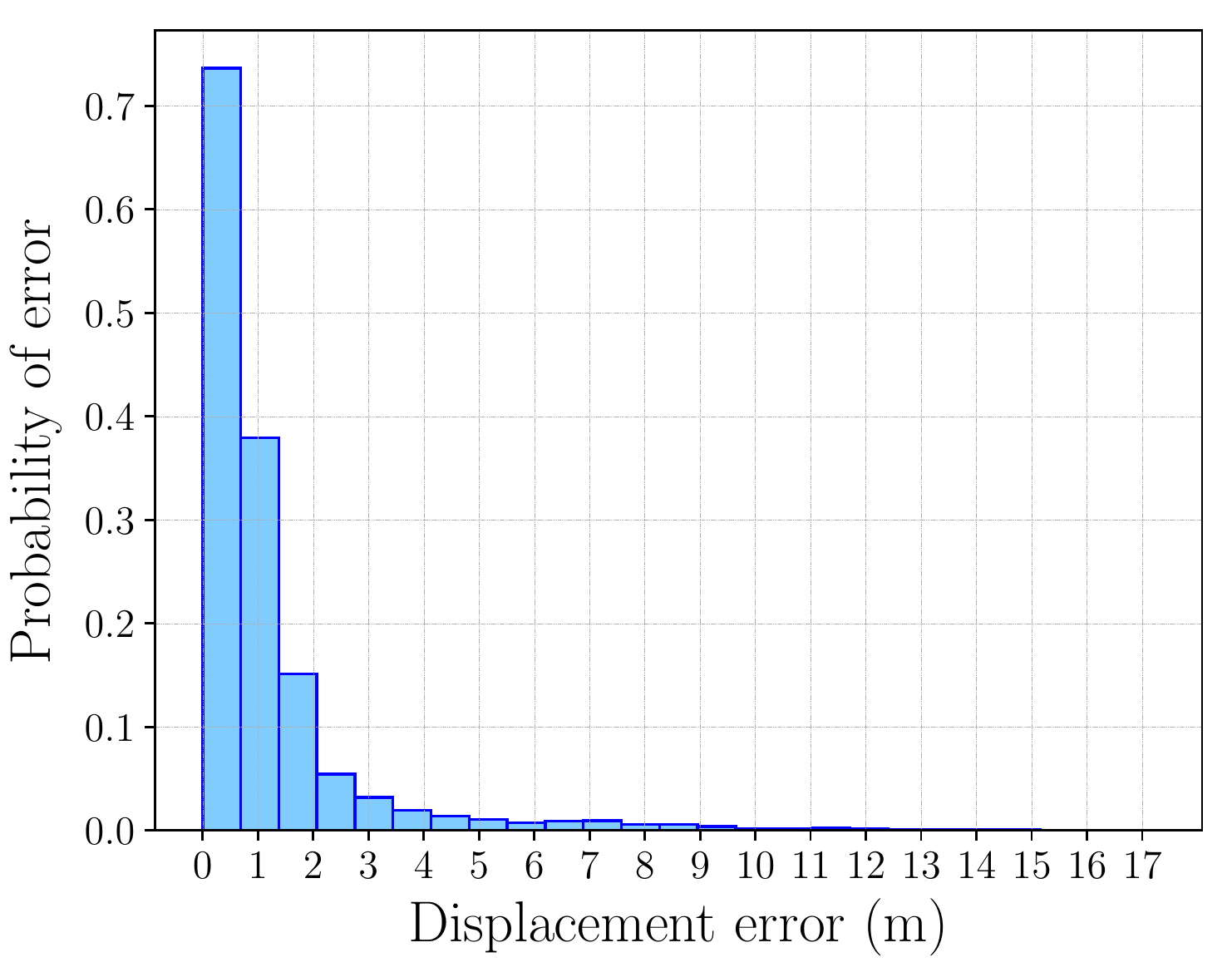}
        \caption{}
        \label{fig:frame_histogram}
    \end{subfigure}
    \caption{(a) The average pixel error vs  number of epochs for a given split. For the first $10$ epochs, we predict one frame into the future. Subsequently, deep future prediction is done and hence there's a sudden rise in loss. (b) Histogram of per frame $L2$-norm for all frames in KITTI, for the \emph{INFER-Skip} model. $86$\% of all our predictions lie within a $2$m threshold.}
\end{figure}

\subsection{Training Details}
All the models were implemented in PyTorch and trained on a single NVIDIA GeForce GTX $1080$ GPU for a maximum of $60$ epochs each. The training process takes about $6$-$8$ hours. The models were trained using the ADAM optimizer with a learning rate of $0.0001$ and gradient clipping with L2 norm of $10.0$. The loss trends across train \& validation sets for \emph{INFER-Skip} \& \emph{ConvLSTM Baseline} is shown in figure \ref{fig:network_loss}. The distribution of L2 norm of all frames across the KITTI dataset is shown in figure \ref{fig:frame_histogram}. 
\begin{table}[!hbt]
	\centering
	
    
    \begin{adjustbox}{max width=\linewidth}
	\begin{tabular}{|c|c|c|c|c|}
		\hline
    	Method &  \textbf{$\mathbf{1}$ sec} & \textbf{$\mathbf{2}$ sec} & \textbf{$\mathbf{3}$ sec} & \textbf{$\mathbf{4}$ sec} \\
      	\hline
        Markov-Baseline & $0.70$ & $1.41$ & $2.12$ & $2.99$\\
      	
      	RNN Encode-Decoder \footnotemark
      	 \cite{virdi2017using} & $0.68$ & $1.94$ & $3.2$ & $4.46$\\
      	
      	ConvLSTM-Baseline (Top 5) & $0.76$ & $1.23$ & $1.60$ & $1.96$\\
      	\hline
      	DESIRE-SI (Best) \cite{desire} & $0.51$ & $1.44$ & $2.76$ & $4.45$\\
      	
      	DESIRE-SI (10\%) \cite{desire} & $\mathbf{0.28}$ & $\mathbf{0.67}$ & $1.22$ & $2.06$\\
      	\hline
      	INFER (Top 5) & $0.61$ & $0.87$ & $1.16$ & $1.53$\\
      	
        INFER-Skip (Top 1) & $0.75$ & $0.95$ & $1.13$ & $1.42$\\
        
        INFER-Skip (Top 3) & $0.63$ & $0.82$ & $1.00$ & $1.30$\\
        
        INFER-Skip (Top 5) & $0.56$ & $0.75$ & $\mathbf{0.93}$ & $\mathbf{1.22}$\\
        \hline

	\end{tabular}
    \end{adjustbox}
    \caption{Quantitative results of baseline models vs. \emph{INFER-Skip} across KITTI dataset for the task of predicting upto $4$s into the future. Error metrics reported are ADE in metres. We have a single model that predicts $1$s, $2$s, $3$s \& $4$s into the future. }
        
    \label{table:quantitative_results}
\end{table}

\footnotetext{\cite{virdi2017using} predicts only upto $2$s into the future, hence we interpolate the values to get predictions for $3$ \& $4$s into the future.}


\subsection{Performance evaluation and transfer results}
\begin{figure}[!b]
    \begin{subfigure}{0.5\linewidth}
        \centering
        \includegraphics[width=\linewidth]{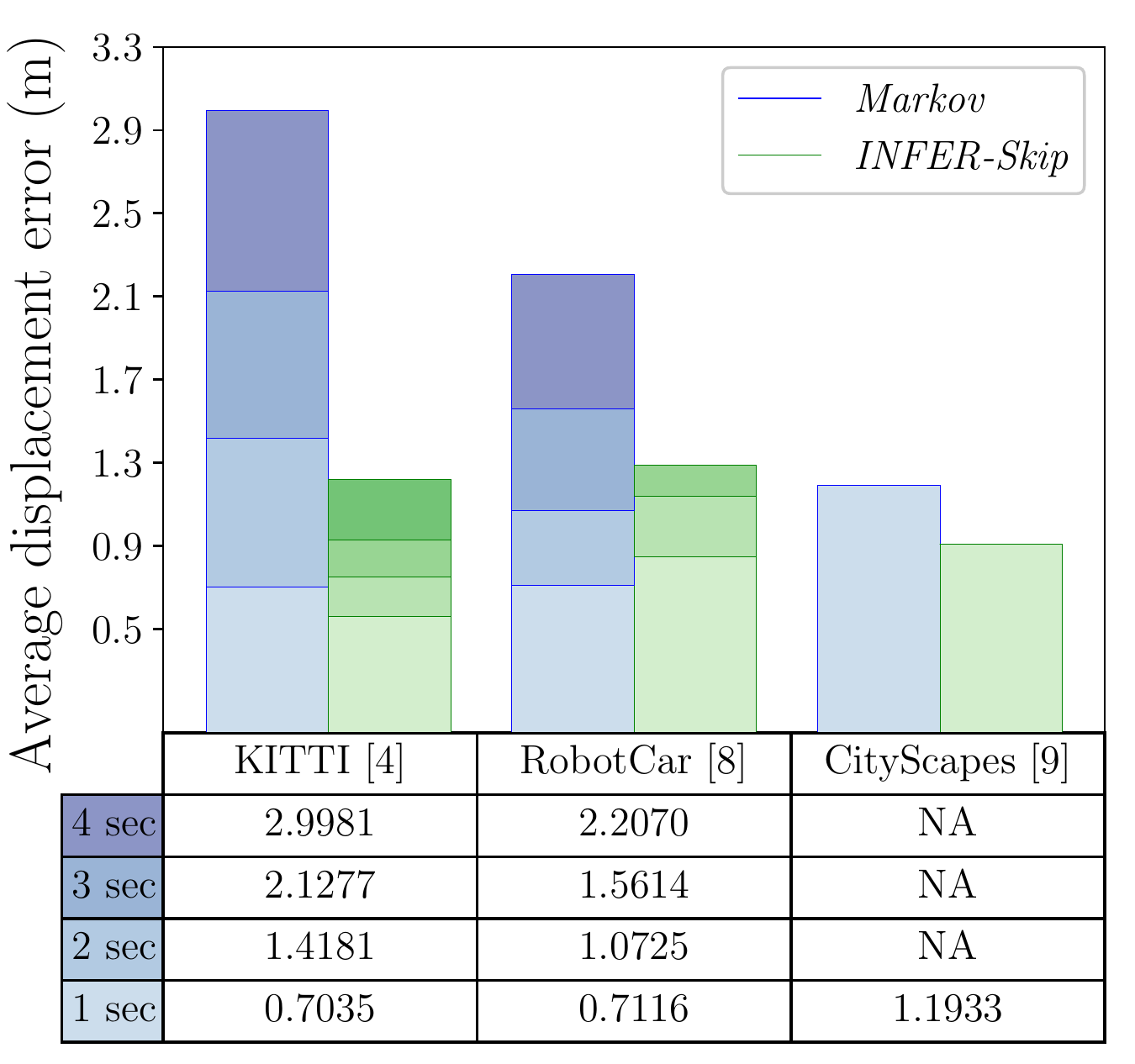}
        \caption{}
        \label{fig:markov_results}
    \end{subfigure}%
    \begin{subfigure}{0.5\linewidth}
        \centering
        \includegraphics[width=\linewidth]{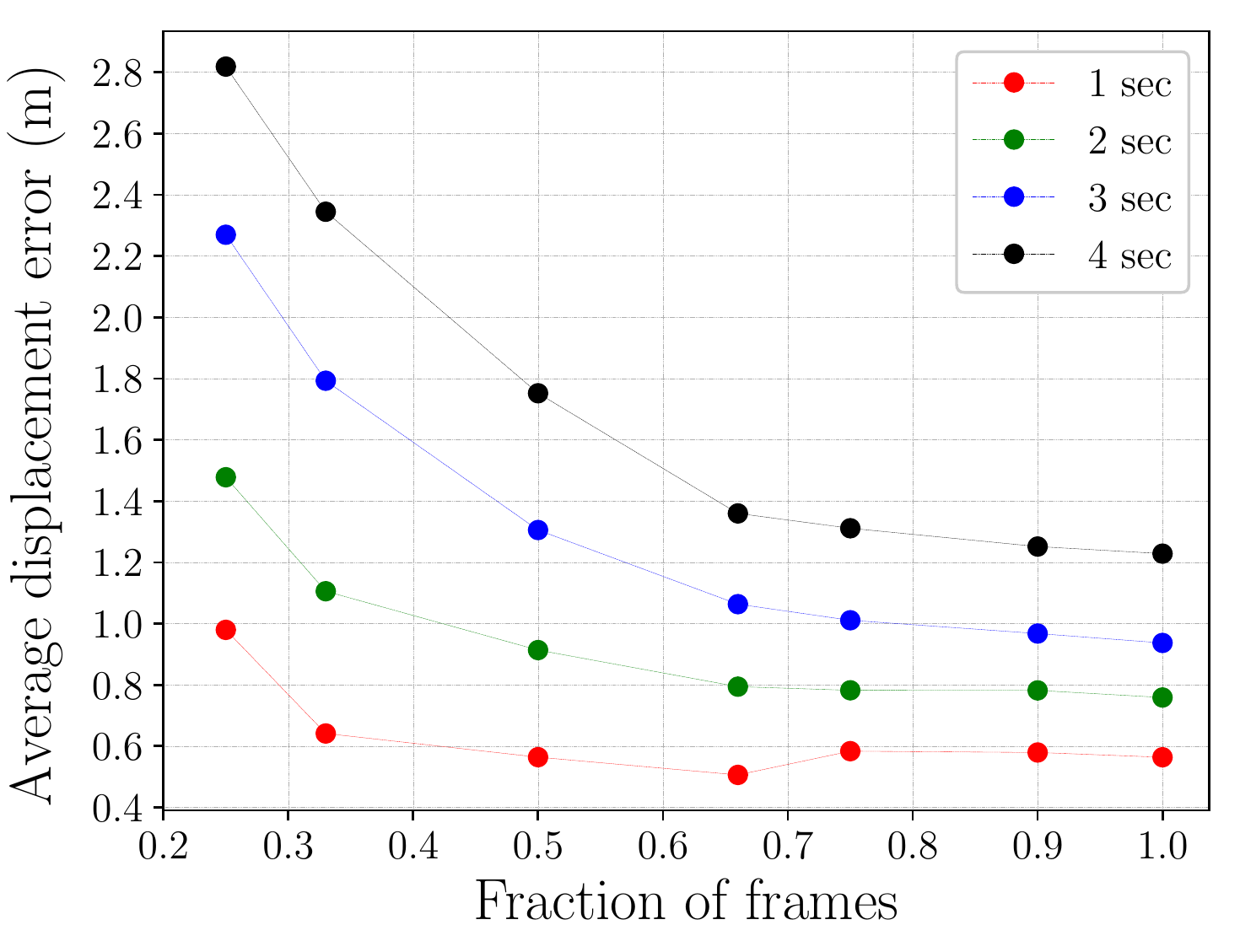}
        \caption{}
        \label{fig:ablation_fps}
    \end{subfigure}
    \caption{(a) Shows the performance of the \emph{Markov baseline} across three different datasets, for predictions upto 4 seconds, and is compared with \emph{INFER-Skip}. (b) Shows the performance of \emph{INFER-Skip} for different frame rates.} 
\end{figure}

The results of the \emph{Markov baseline} are shown in Fig.  \ref{fig:markov_results}. It performs significantly better than the \emph{RNN Encoder Decoder} \cite{virdi2017using} due to the use of grid based representation for the target vehicle. Note that  both these approaches output a single prediction. The \emph{ConvLSTM} baseline outperforms these two while predicting deep into the future, while degrading slightly when predicting upto $1$s into the future.

Our model \emph{INFER-Skip} which consists of convolutional LSTMs along skip connections outperforms current state of the art  DESIRE \cite{desire} in KITTI \cite{KITTI} while predicting deep into the future($3$s,$4$s). It outperforms all other models on all the evaluation metrics.
\emph{INFER} which consists of only a single convolutional LSTM also beats the current state of the art method \cite{desire} for predictions deep into the future ($3$s \& $4$s). These models do not perform as well for shorter timesteps in the future as we are limited by the resolution of the intermediate representation, hence making the comparison favourable for \cite{desire}.  Being multi-modal in nature, we can predict multiple future trajectories for a given track history by sampling the top $K$ samples at each time step. We report the performance of \emph{INFER-Skip} with several values of $K=1$, $3$ \& $5$ respectively. Using the top $5$ predictions significantly improves performance by upto $20$ cms, demonstrating the multi-modal nature of the model.

\begin{table}[!tb]
	\centering
    \begin{adjustbox}{max width=\linewidth}
	\begin{tabular}{|c|c|c|}
		\hline
    	Method &  \textbf{$\mathbf{1}$ sec (\text{*})} & \textbf{$\mathbf{1}$ sec (\text{**})} \\
      	\hline
      	ConvLSTM-Baseline (Top 1) & $1.5$ & $1.23$ \\
      	
      	ConvLSTM-Baseline (Top 3) & $1.36$ & $1.09$ \\
      	
      	ConvLSTM-Baseline (Top 5) & $1.28$ & $1.021$ \\      	
      	
      	\hline
      	INFER-skip (Top 1) & $1.11$ & $1.12$ \\
      	
      	INFER-skip (Top 3) & $0.99$ & $0.98$ \\
      	
      	INFER-skip (Top 5) & $\mathbf{0.91}$ & $\mathbf{0.91}$ \\
        \hline
	\end{tabular}
    \end{adjustbox}
    \caption{Transfer results of \emph{INFER-Skip} \& the \emph{ConvLSTM Baseline} models on Cityscapes \cite{cityscapes}. We report the ADE in metres.}
    \label{table:cityscapes_transfer}
\end{table}
\begin{table}[!hbt]
	\centering
    \begin{adjustbox}{max width=\linewidth}
	\begin{tabular}{|c|c|c|c|c|}
		\hline
    	Method &  \textbf{$\mathbf{1}$ sec} & \textbf{$\mathbf{2}$ sec} & \textbf{$\mathbf{3}$ sec} & \textbf{$\mathbf{4}$ sec} \\
      	\hline
        INFER-Skip & $0.85$ & $1.14$ & $1.29$ & $1.50$\\
        \hline
	\end{tabular}
    \end{adjustbox}
    \caption{Transfer results of \emph{INFER-Skip} model tested on a few sequences of Oxford Robotcar. We report the ADE in metres.}
    \label{table:oxford_car}
\end{table}
\begin{table}[!hbt]
	\centering
    \begin{adjustbox}{max width=\linewidth}
	\begin{tabular}{|c||c|c|c|c||c|}
		\hline
		\multirow{2}{*}{Method}
		    & \multicolumn{4}{c||}{KITTI}
		        & \multicolumn{1}{c|}{Cityscapes} \\ \cline{2-6}
        &  \textbf{1 sec} & \textbf{2 sec } & \textbf{3 sec} & \textbf{4 sec} & \textbf{1 sec} \\ \hline
        INFER-Skip  & $0.56$ & $\mathbf{0.75}$ & $\mathbf{0.93}$ & $1.22$ & $\mathbf{1.12}$  \\
      	
        INFER-Skip w/o road  & $0.70$ & $1.20$ & $1.80$ & $2.49$ & $5.62$\\
        
        INFER-Skip w/o obstacles  & $\mathbf{0.54}$ & $0.80$ & $1.00$ & $1.24$ & $1.21$\\
        
        INFER-Skip w/o lane  & $0.57$ & $0.76$ & $0.94$ & $\mathbf{1.21}$ & $1.33$\\
        
        \hline
	\end{tabular}
    \end{adjustbox}
    \caption{Ablation results across KITTI \& Cityscapes by removing semantics corresponding to road, lane \& obstacles from the intermediate representation. We report the ADE in metres.}
    \label{table:ablationKITTICityscapes}
\end{table}

\begin{figure*}[!t]
    \begin{subfigure}{0.25\textwidth}
        \centering
        \includegraphics[width=\linewidth]{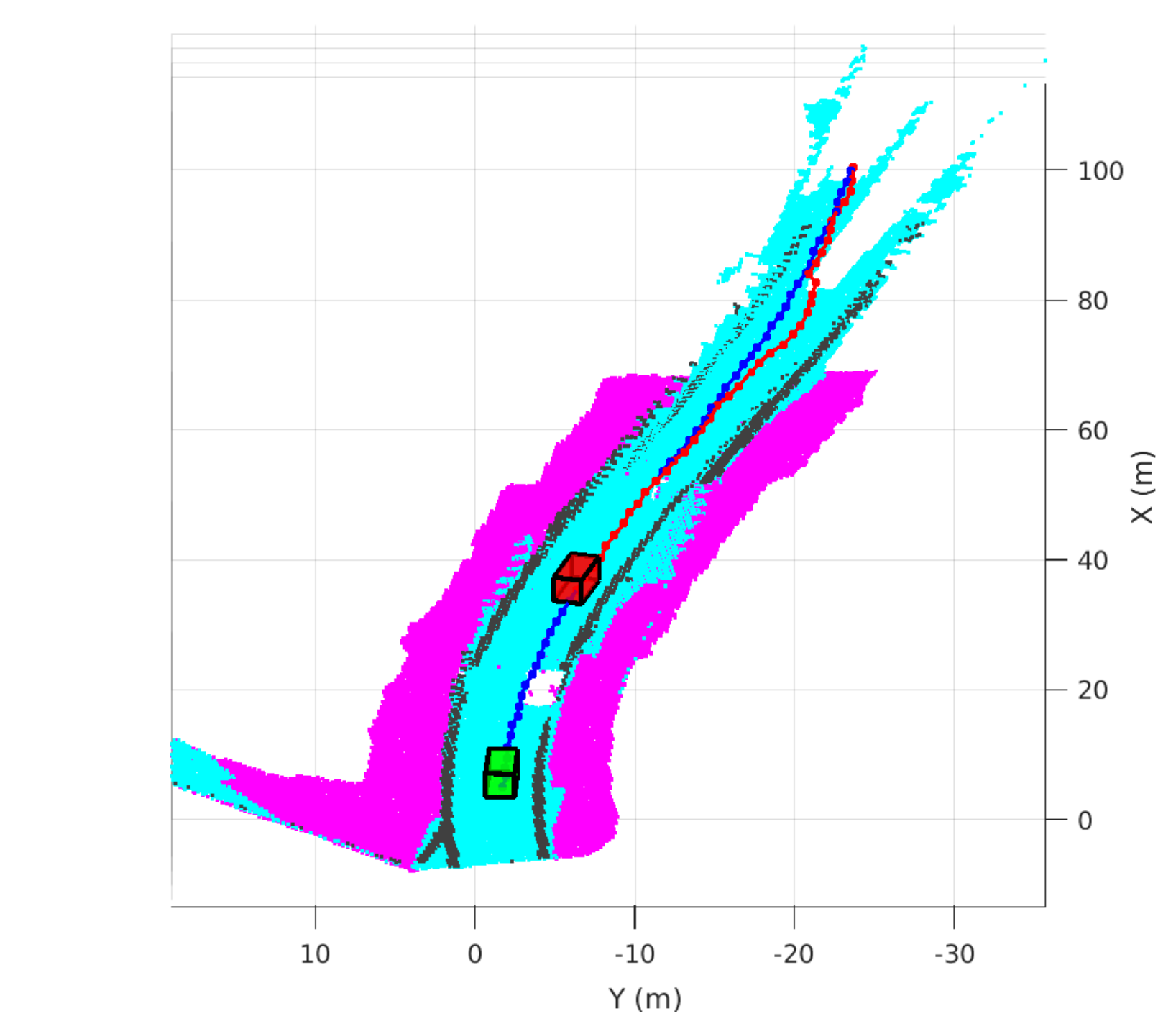}
    \end{subfigure}%
    \begin{subfigure}{0.25\textwidth}
        \centering
        \includegraphics[width=0.85\linewidth]{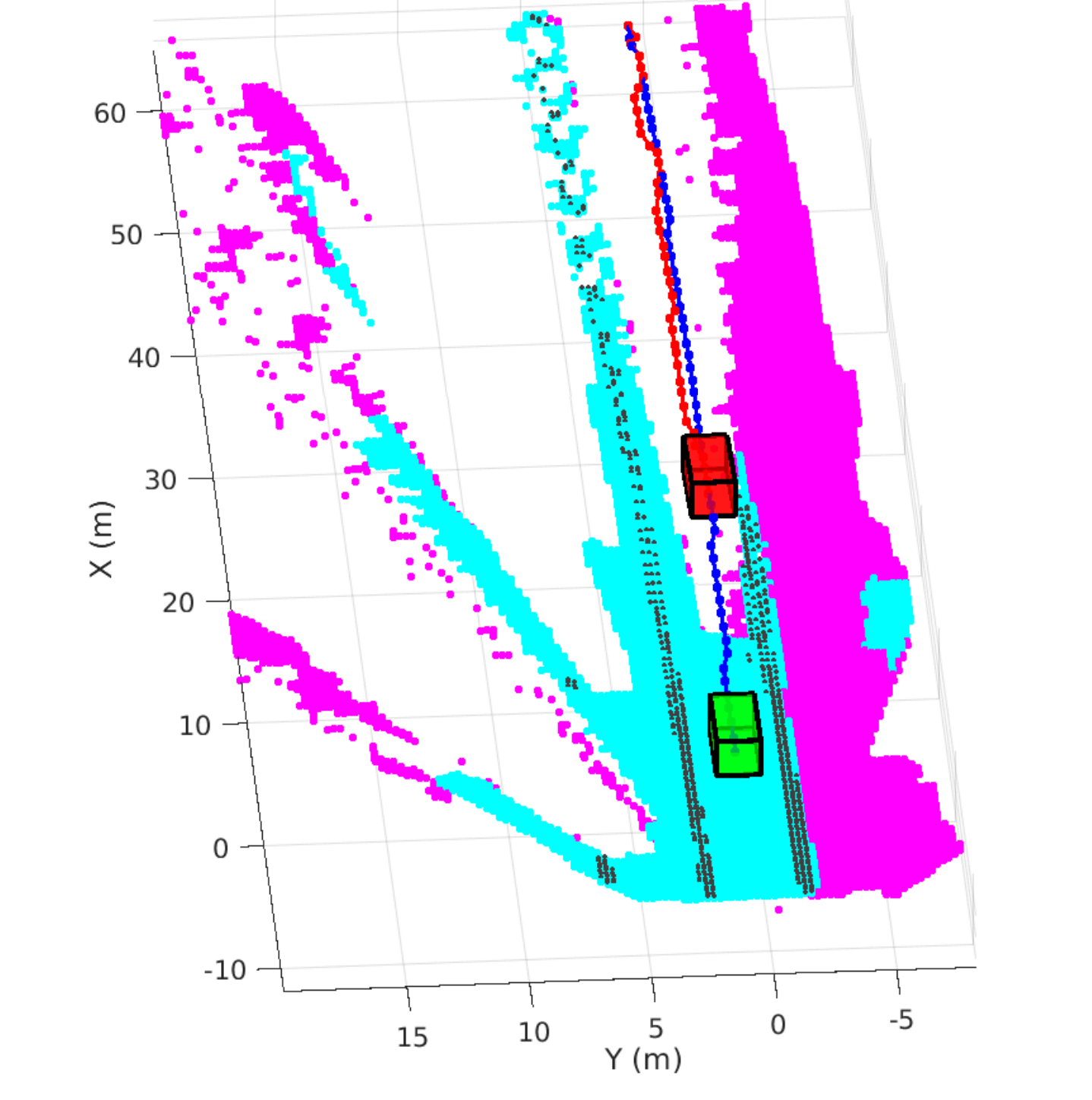}
    \end{subfigure}%
    \begin{subfigure}{0.25\textwidth}
        \centering
        \includegraphics[width=\linewidth]{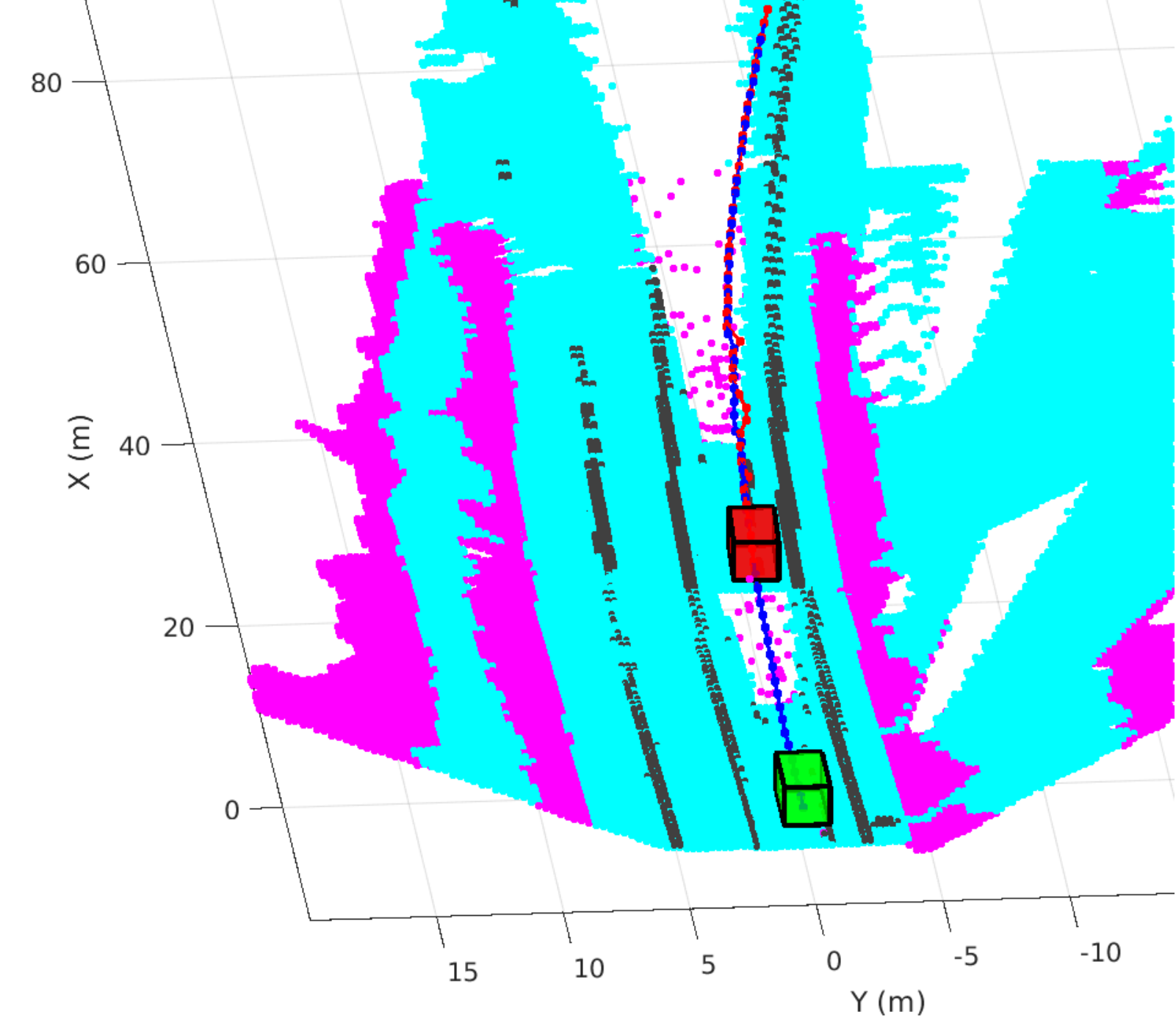}
    \end{subfigure}%
    \begin{subfigure}{0.25\textwidth}
        \centering
        \includegraphics[width=0.94\linewidth]{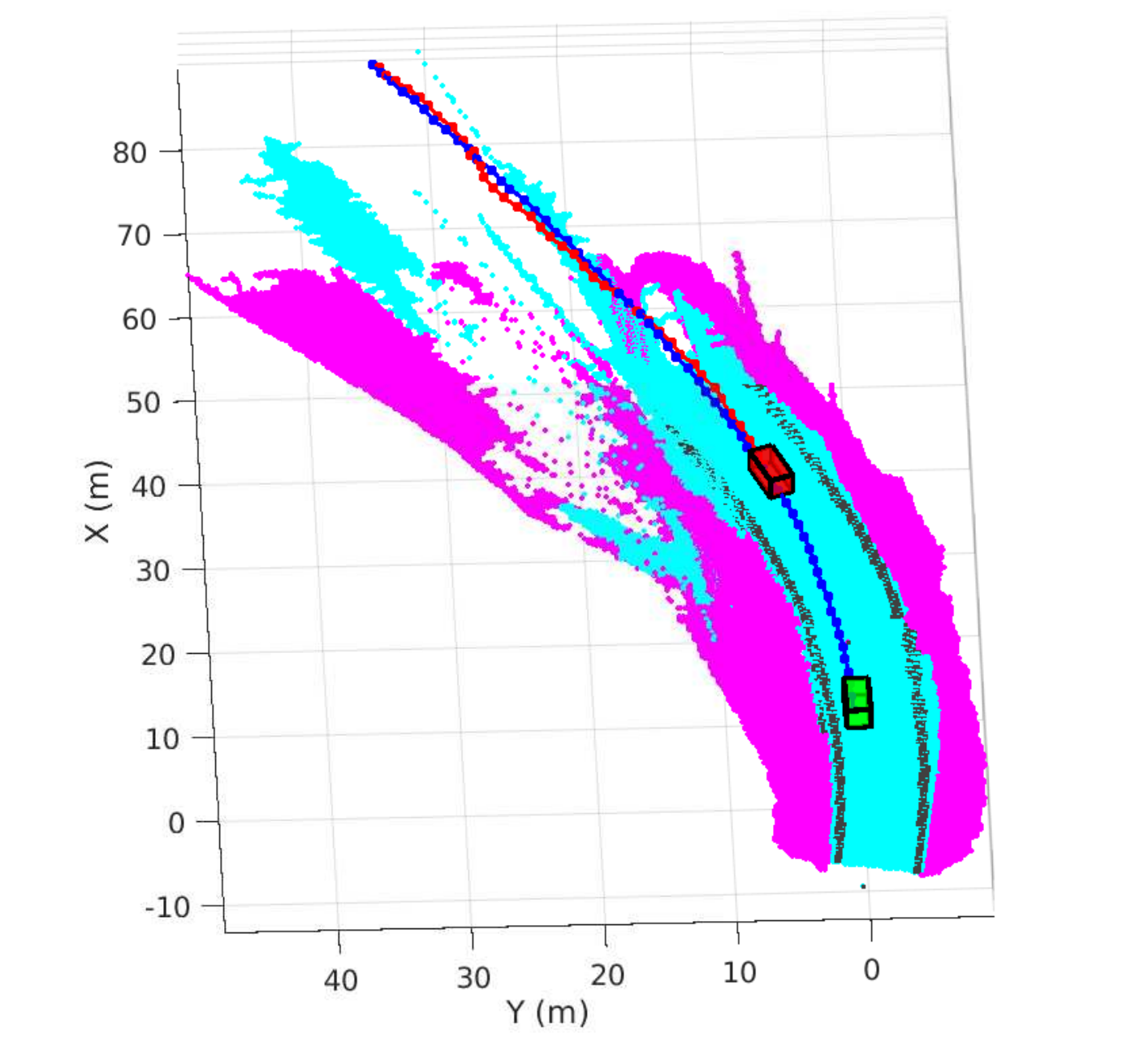}
    \end{subfigure}%
    \caption{The above qualitative results from the validation fold of KITTI \cite{KITTI} showcase the efficacy of \emph{INFER-Skip} in using the intermediate representation to predict complex trajectories. For example, in the left most plot, the network is able to accurately predict the unseen second curve in the trajectory (predicted and ground truth trajectories are shown in red and blue color, respectively). The green and red 3D bounding boxes indicate start of preconditioning and start of prediction of the vehicle of interest, respectively. It is worth noting that the predicted trajectories are well within the lane (dark gray) and road region (cyan), while avoiding collisions with the obstacles (magenta).}
    \label{fig:qualitative}
\end{figure*}

\textbf{Transfers: } We test the transfer ability of our model by training on KITTI \cite{KITTI} \& testing on Cityscapes \& Oxford Robotcar dataset \cite{cityscapes, oxfordCar}. We precondition our model for a total of $0.8$s and test upto $1$s into the future for Cityscapes dataset \cite{cityscapes} as it provides only sequences of length $1.76s$. We present the performance of the \emph{INFER-Skip} across Cityscapes in Table \ref{table:cityscapes_transfer}. In (\text{*}), the results of \emph{INFER-Skip} without vehicle channel are shown. This model transfers well to the Cityscapes dataset which differs a lot from the KITTI dataset in terms of scene layout, weather condition \& vehicle motion. The performance accross KITTI \& Cityscapes is off by only $34$ cm for $1$s into the future. We find \emph{ConvLSTM-Baseline} trasfers well to the Cityscapes \cite{cityscapes} dataset which suggests that using a birds eye view helps in transfer, as such a representation does not rely heavily on the camera viewing angle and camera mounting parameters. However, \emph{INFER-Skip} performs better than the \emph{ConvLSTM Baseline} highlighting the potent of intermediate representations over RGB data. All our models were trained with a preconditioning of $2$s and then predict upto $4$s into the future. Thus, predicting the future trajectories in Cityscapes with $1$s preconditioning leads to higher than expected error. In (\text{**}), we train our main model \emph{INFER-Skip} \& the  \emph{ConvLSTM Baseline} on another split of the KITTI dataset that consists of trajectories of $2$s length only. These models were trained with only $1$s preconditioning and the results of their transfer on Cityscapes dataset is shown in the $2$nd column of Table \ref{table:cityscapes_transfer}.

We further test transfer of \emph{INFER-Skip} on a few trajectories of the Oxford Robot Car \cite{oxfordCar} dataset and the performance does not degrade while predicting deep into the future as can be seen in the table \ref{table:oxford_car}. Our method is able to generalize well to drastic change in scene layout (Germany to UK) and right-side to lefts-side driving.

\subsection{Ablation study}
We conduct detailed ablation studies on \emph{INFER-Skip} to determine the type of semantics that help the model perform well across KITTI\cite{KITTI} and Cityscapes\cite{cityscapes}. For \cite{cityscapes}, we directly do ablation on the transfer rather than training on it first.  We experiment with 3 major variants of \emph{INFER-Skip} with the road, obstacles \& lane channels removed. The results in Table \ref{table:ablationKITTICityscapes} show that the scene semantics do indeed play a vital role in the performance of our proposed model. The removal of the road channel reduces the performance of the model drastically in KITTI. The performance of the model degraded severely when tested on Cityscapes as the error reaches as high as $5.62$m. While the other channels like obstacles \& lane do not seem to affect the performance of the model in the KITTI dataset greatly, the corresponding transfer error in Cityscapes increases significantly, showcasing how semantics play a critical role in transferring across datasets. Thus, our representations are \emph{representationally adequate} and \emph{generalizes} well to other datasets. We also show ablation on the frame rate in Fig. \ref{fig:ablation_fps} varying it while testing on KITTI\cite{KITTI}. We show the loss trend for 1, 2, 3, and 4 seconds into the future. The variation in frame rate captures the change in relative velocity of VoI w.r.t ego vehicle. We observe that even when the frame rate is dropped to $60\%$, the performance does not degrade significantly, highlighting the potent of the representations to generalize to different relative velocities. 

\subsection{Qualitative results}
\begin{figure}[!t]
    \begin{subfigure}{0.25\textwidth}
        \centering
        \includegraphics[width=0.92\linewidth]{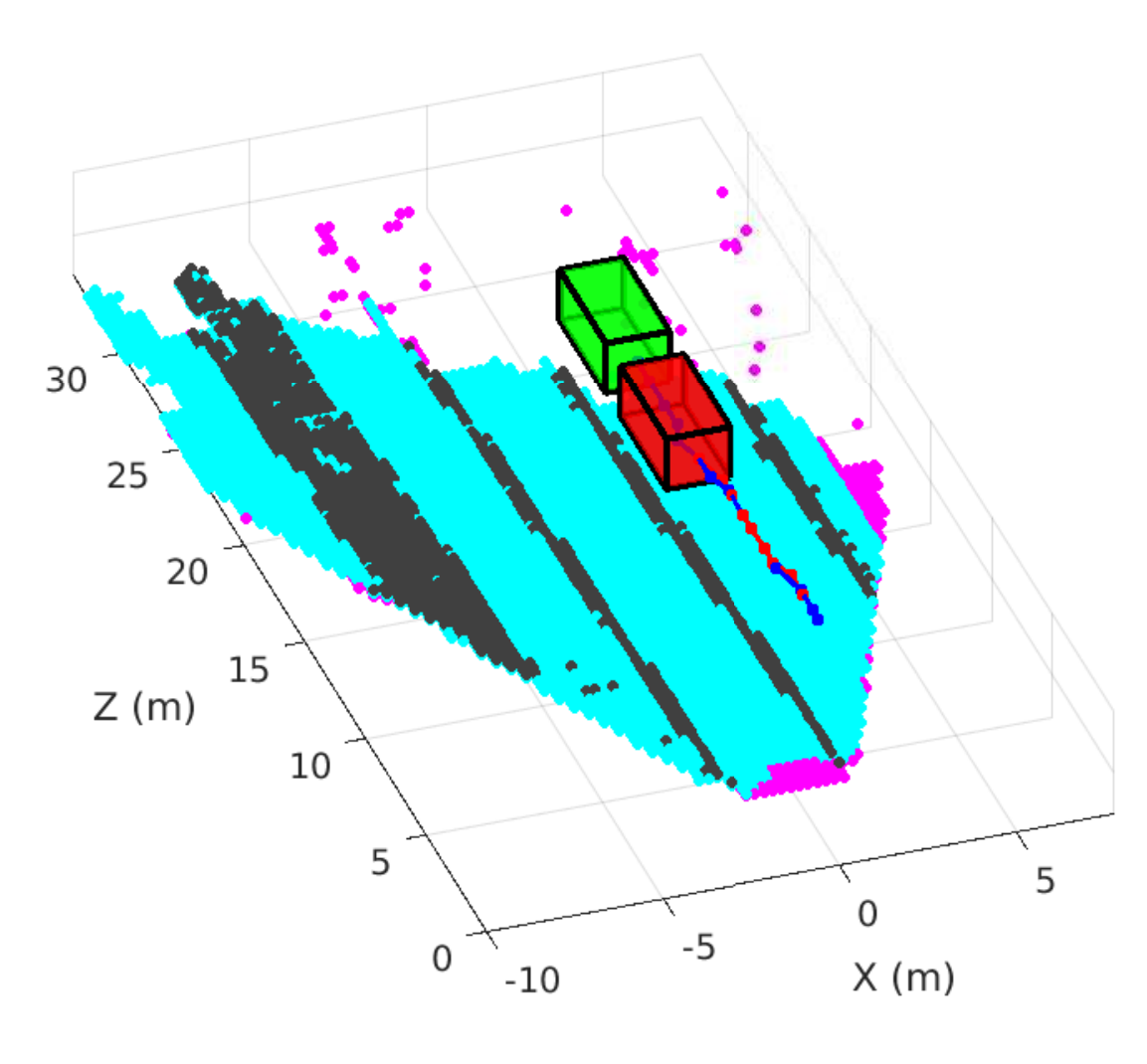}
        \caption{}
        \label{fig:qualitativeCityscapes}
    \end{subfigure}%
    \begin{subfigure}{0.25\textwidth}
        \centering
        \includegraphics[width=0.94\linewidth]{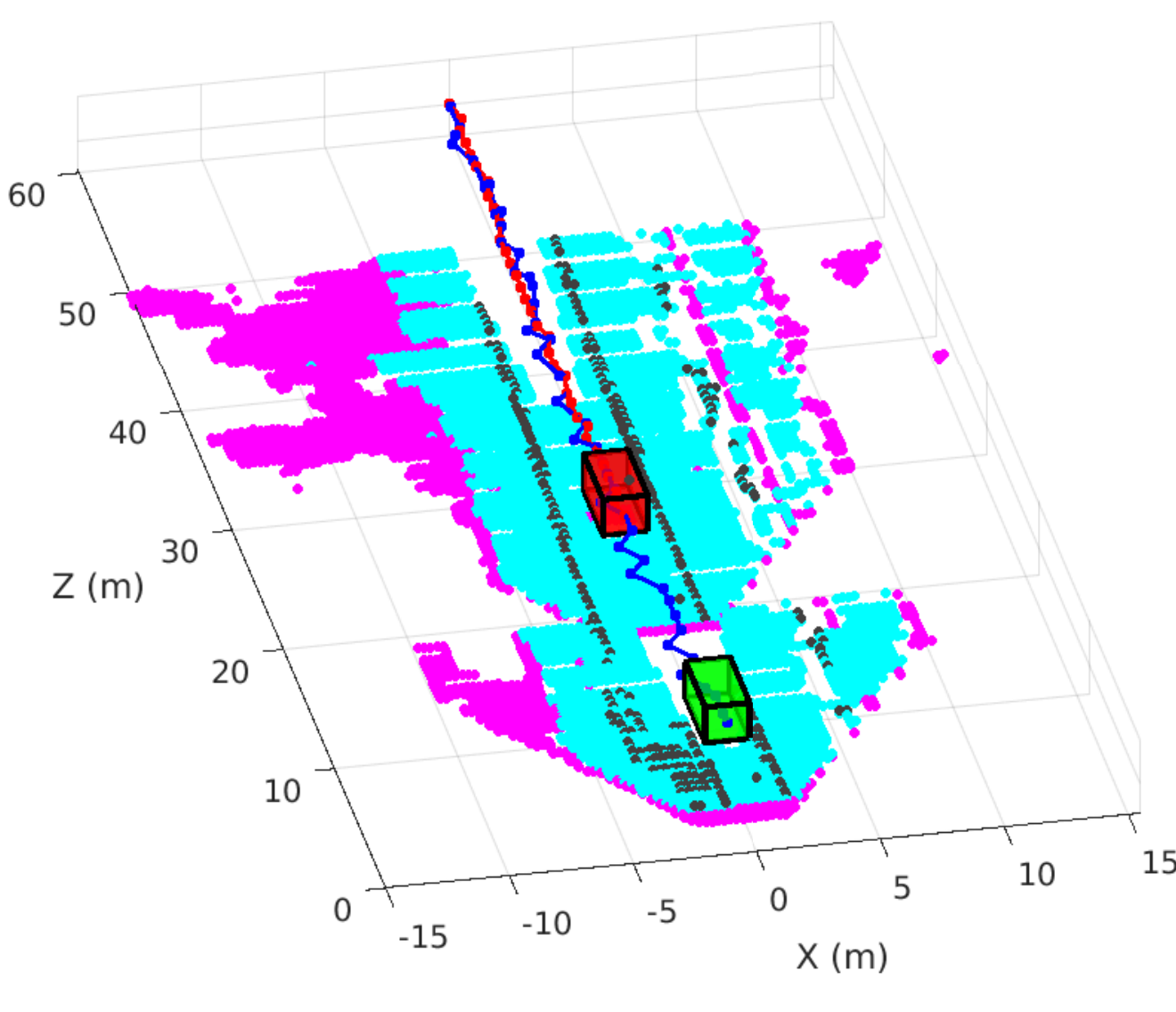}
        \caption{}
        \label{fig:qualitativeOxford}
    \end{subfigure}%
    \caption{Above figure qualitatively shows the result of zero-short transfer of \textit{INFER-Skip}, trained on KITTI dataset \cite{KITTI}, to (a) Cityscapes  \cite{cityscapes} and (b) Oxford robot car \cite{oxfordCar} datasets. The color conventions are same as that of the plots in Fig. 
    \ref{fig:qualitative}.}
\end{figure}

We showcase some qualitative results of our approach on challenging KITTI \cite{KITTI}, Cityscapes \cite{cityscapes}, and Oxford Robotcar \cite{oxfordCar} sequences in figures \ref{fig:qualitative},  \ref{fig:qualitativeCityscapes} \& \ref{fig:qualitativeOxford} respectively. These results illustrate the effectiveness of our \emph{intermediate representations} and models to predict complex trajectories and transfer zero-shot across datasets \cite{oxfordCar,cityscapes}. The green 3D bounding box depicts the first sighting of the vehicle of interest which is also when we start preconditioning, and the red 3D bounding box indicates the start of prediction. The plots clearly show that \emph{INFER-Skip}, using the proposed \emph{intermediate representations}, is accurately predicting the trajectories. It can also be seen that even non-trivial trajectories constituting of multiple turns are predicted very close to their respective ground truth while being within the road (cyan) and lane (dark gray) regions; obstacles are shown in magenta color. For the purpose of visualizaition, all the trajectories along with the intermediate representations (road, lane, and obstacle) are registered in the ego vehicle coordinate frame corresponding to the start of prediction.

\begin{table}[!hbt]
	\centering
    \begin{adjustbox}{max width=\linewidth}
	\begin{tabular}{|c|c|c|}
		\hline
    	&  \textbf{KITTI \cite{KITTI}} & \textbf{Cityscapes \cite{cityscapes}}  \\
    	\hline
        Association accuracy & $85.71\%$ & $75\%$  \\
        \hline
	\end{tabular}
    \end{adjustbox}
    \caption{Object association for Multi-Object tracking}
    \label{table:object_tracking}
\end{table}
\subsection{Summary of results}
The cornerstone of this effort is that intermediate representations are well apt for the task of future prediction. In Table \ref{table:quantitative_results} we highlight that using semantics provides a significant  boost over techniques that operate over raw pixel intensities/disparities. In Table \ref{table:cityscapes_transfer}, \ref{table:oxford_car} we showcase the efficacy of intermediate representations to transfer across different datasets collected across different cities. We perform an extensive ablation study on which semantics help in transfer and report the results of this ablation on KITTI \cite{KITTI} and Cityscapes \cite{cityscapes} in Table \ref{table:ablationKITTICityscapes}, highlighting how semantics are critical in transfer across dataset. We do an ablation study on frame rates in Fig. \ref{fig:ablation_fps} and showcase the generalizabilty of our approach to the change in relative velocity of VoI and ego vehicle. We highlight a usecase of our approach for object association in multi-object tracking based on future predicted locations in Table \ref{table:object_tracking}. Association is done based on the minimum $L2$ distance the future location of VoI has with all the vehicles present in the scene. We showcase comparable performance on \cite{KITTI,cityscapes} to popular approaches like \cite{eccv2016}. 

\section{Conclusions}
\label{sec:conclusions}
We propose \emph{intermediate representations} that are apt for the task of future trajectory prediction of vehicles. As opposed to using raw sensor data, we condition on \emph{semantics} and train an autoregressive network to accurately predict future trajectories of vehicles. We outperform the current state of the art approaches, demonstrating that semantics provide a significant boost over techniques that operate solely over raw pixel intensities or depth information. We show that our representations and models transfer zero-shot to completely different datasets, collected across different cities, weather conditions, and driving scenarios. We carry out a thorough ablation study on the importance of our semantics and show generatlization of our approach on different frame rates. Additionally, we demonstrate an application of our approach in data association in multi-object tracking.



\bibliography{00root}
\bibliographystyle{IEEEtran}

\end{document}